\documentclass[runningheads]{llncs}
\usepackage{graphicx}
\usepackage{natbib}
\usepackage{amsmath}

\usepackage{url}

\begin{document}
\title{Optimising Knee Injury Detection with Spatial Attention and Validating Localisation Ability}
\titlerunning{Knee Injury Detection and Validating Localisation Ability}
%
%
\author{Niamh Belton\inst{1,2}\orcidID{0000-0003-4949-4745} \and
Ivan Welaratne\inst{3}\orcidID{0000-0003-1376-5259} \and
Adil Dahlan\inst{2}\orcidID{0000-0003-1457-8228} \and
Ronan T Hearne\inst{4}\orcidID{0000-0003-0248-6908} \and
Misgina Tsighe Hagos\inst{1,5}\orcidID{0000-0002-9318-9417} \and
Aonghus Lawlor\inst{5,6}\orcidID{0000-0002-6160-4639} \and
Kathleen M. Curran\inst{1,2}\orcidID{0000-0003-0095-9337}}
\authorrunning{N. Belton et al.}
%
\institute{Science Foundation Ireland Centre for Research Training in Machine Learning \and School of Medicine, University College Dublin  \and
Department of Radiology, Mater Misericordiae University Hospital, Dublin \and School of Electronic Engineering, University College Dublin \and School of Computer Science, University College Dublin\and 
Insight Centre for Data Analytics, University College Dublin, Dublin, Ireland
\email{\{niamh.belton, ronan.hearne, misgina.hagos\}@ucdconnect.ie \\ \{adil.dahlan, aonghus.lawlor, kathleen.curran\}@ucd.ie}
}
\maketitle              
\vspace{-0.2cm}
\begin{abstract}
This work employs a pre-trained, multi-view Convolutional Neural Network (CNN) with a spatial attention block to optimise knee injury detection. An open-source Magnetic Resonance Imaging (MRI) data set with image-level labels was leveraged for this analysis. As MRI data is acquired from three planes, we compare our technique using data from a single-plane and multiple planes (multi-plane). For multi-plane, we investigate various methods of fusing the planes in the network. This analysis resulted in the novel \lq{}MPFuseNet\rq{} network and state-of-the-art Area Under the Curve (AUC) scores for detecting Anterior Cruciate Ligament (ACL) tears and Abnormal MRIs, achieving AUC scores of 0.977 and 0.957 respectively. We then developed an objective metric, Penalised Localisation Accuracy ($PLA$), to validate the model's localisation ability. This metric compares binary masks generated from Grad-Cam output and the radiologist's annotations on a sample of MRIs. We also extracted explainability features in a model-agnostic approach that were then verified as clinically relevant by the radiologist.

\vspace{-0.3cm}
\end{abstract}
\section{Introduction}
\vspace{-0.2cm}
Knee injuries are one of the most prevalent injuries in sporting activities \citep{kennedy1993sports}. Musculoskeletal (MSK) knee injuries can be detrimental to athletes' careers. Such injuries require early intervention and appropriate rehabilitation. Magnetic Resonance Imaging (MRI) is the gold standard imaging modality for non-invasive knee injury diagnosis \citep{kneemri}. MRI is a volumetric imaging technique that is acquired from three planes, namely axial, coronal and sagittal. Machine Learning (ML) can be used to develop Computer Aided Detection (CAD) systems that automate tasks such as detecting injuries from MRIs. These systems have several benefits including reduced diagnosis time for patients and alleviating the ever-growing workload of radiologists by assisting diagnosis.

With the growing interest in CAD systems, there is a corresponding growing requirement for the CAD systems to be explainable.  Explainability increases clinicians' confidence in the models and allows for smoother integration of these models into clinical workflows \citep{floruss2020artificial}. Validating the localisation ability and extracting features of automated injury detection systems aids their explainability. There has been significant research in the area of assessing the localisation ability of various saliency techniques \citep{untrust,hist}. There is, however, a lesser focus on assessing a model's localisation ability. While object detection algorithms can be specifically trained to locate the site of the target task, they have increased model complexity and they require annotations of the complete data set. It is more often that saliency maps are employed to assess the localisation ability of a classification model. In this work, we validate the model's localisation ability by comparing the saliency map technique Grad-Cam \citep{gradcam} to annotations defined by the radiologist. This analysis found that existing segmentation metrics such as Intersection over Union (IoU) and Dice Coefficient are not suitable for quantifying the localisation ability of a model. This highlights the need for an interpretable metric that communicates an accurate representation of a model's localisation ability.

In this work, we propose a CNN with a ResNet18 architecture and integrated spatial attention block for MSK knee injury detection. The main aspects of this work are as follows.
\renewcommand{\labelenumii}{\Roman{enumii}}
 \begin{enumerate}
 
 \item We investigate if an injury can be accurately detected using data from a single-plane or if additional planes are required for optimal detection accuracy.
 \item For cases where additional planes increase detection accuracy, we investigate methods of fusing planes in a network and develop a new multi-plane network, \lq{}MPFuseNet\rq{}, that achieves state-of-the-art performance for ACL tear detection. 
 \item We develop an objective metric, Penalised Localisation Accuracy ($PLA$), to validate the localisation ability of the model. This proposed metric communicates a more accurate representation of the model's localisation ability than typical segmentation metrics such as IoU and Dice. 
 \item We extract explainability features using a post-hoc approach. These features were then validated as clinically relevant.
 \end{enumerate}

\vspace{-0.4cm}
\section{Related Work}
\vspace{-0.2cm}
The research in the field of ML with MRI for MSK injury detection was accelerated by the release of the open-source MRNet data set \citep{mrnet}. This is a rich data set of knee MRIs with image-level labels acquired from the Stanford Medical Centre. \cite{mrnet} implemented a pre-trained AlexNet architecture for each plane and trained separate models for detecting different types of injuries. \cite{dcu} followed a similar approach and employed a ResNet18 with advanced data augmentation, while \cite{irmakci2019deep} investigated other classic Deep Learning (DL) architectures. \cite{elnet} employed the ELNet, a CNN trained end-to-end with additional techniques such as multi-slice normalisation and blurpool. This method was shown to have superior performance than previous methods. Other research studies have also demonstrated the effectiveness of using Deep Learning for detecting ACL tears on additional data sets \citep{ref1}. 

Spatial attention is another area that has sparked interest in the ML and medical imaging community. Spatial attention is a technique that focuses on the most influential structures in the medical image. This technique has been shown to improve the performance of various medical imaging tasks such as image segmentation \citep{unet} and object detection \citep{nvidia}. \cite{nvidia} implemented 3D spatial and contextual attention for deep lesion detection. We base our spatial attention block on this work. 

The rapid growth of ML techniques for medical tasks has highlighted the requirement for DL models to be explainable. Several saliency map techniques have been proposed in recent years \citep{gradcam, gradplus, smoothgrad}. These techniques highlight regions of an image that are influential in the output of the model. There has also been extensive work in the area of using ML to extract features from models. Recently, \cite{adri} used Testing with Concept Activation Vectors (TCAV), originally proposed by \cite{TC}, to extract underlying features for cardiac MRI segmentation. These features not only aid the explainability of the model but they can also discover previously unknown aspects of the pathology.

\vspace{-0.2cm}
\section{Materials}
\vspace{-0.3cm}
The MRNet data set first published by \cite{mrnet} consists of 1,250 MRIs of the knee. Each MRI is labelled as having an Anterior Cruciate Ligament (ACL) tear, meniscus tear and/or being abnormal. The MRIs were acquired using varying MRI protocols as outlined in Table S1 in the original paper \citep{mrnet}. Each plane was made available in a different MRI sequence. The  sequences for the sagittal, coronal and axial planes are T2-weighted, T1-weighted and Proton Density-weighted respectively \citep{mrnet}. The MRI data was previously pre-processed to standardise the pixel intensity scale. This was conducted using a histogram intensity technique \citep{mrnet, nyul} to correct inhomogeneous pixel intensities across MRIs. 
\vspace{-0.1cm}
\section{Method}
\vspace{-0.2cm}
\subsection{Model Backbone}
\vspace{-0.2cm}
The backbone model is a 2D multi-view ResNet18. \cite{dcu} demonstrated that a ResNet18 \citep{resnet}, pre-trained on the ImageNet data set \citep{imagenet} outperforms other classic architectures such as AlexNet for detecting knee injuries from MRI data. For this reason, we employ a ResNet18 architecture. The weights were initialised with the pre-trained model weights and all weights were subsequently fine-tuned. 

MRI data is volumetric and therefore, each plane of an MRI is made up of several images known as \lq slices\rq. A multi-view technique was employed to combine slices of the same MRI in the network. \lq{}Views\rq{} in the multi-view network are equivalent to slices. The batch dimension of the multi-view CNN is equal to the number of slices in one MRI. The slices are input into the multi-view CNN and the output has dimensions $(b, f)$ where $b$ is the batch dimension, equal to the number of slices of an MRI and $f$ is a one-dimensional (1D) vector that is representative of each slice. \cite{multiview} demonstrated that an element-wise maximum operation was optimal for combining views. For this reason, we implemented an element-wise maximum operation across the batch dimension to obtain a final 1D vector of combined views (slices) of size $(1, f)$ that is representative of a stack of slices from one plane of an MRI. This vector is then passed through a fully connected layer to produce the final output.

\vspace{-0.3cm}
\subsection{Spatial Attention}
\vspace{-0.1cm}
A soft spatial attention mechanism is integrated into our base model. This follows the spatial attention method proposed by \cite{nvidia} who integrated spatial attention into a VGG16 for deep lesion detection. We opt for a ResNet18 as its performance for this task has been demonstrated and it has approximately 127 million less parameters than the VGG16. The attention block is designed as follows. 

We define our output volume $D^l$ from a convolutional layer $l$ of size $(b, c, h, w)$ where $b$ is number of slices, $c$ is the number of channels (feature maps), and $h$ and $w$ are the height and width of the feature maps. In the attention block, volume $D^l$ is input into a $1 \times 1$ convolution. This produces an output volume A$^l$ of shape $(b, c, h, w)$, where the dimensions are of identical size to D$^l$. A softmax is applied to each feature map of $A^l$ so that each of the $c$ feature maps of dimensions $(h, w)$ sum to one. Small values are sensitive to the learning rate \citep{nvidia} and therefore, each feature map is normalised by dividing the values of the feature map by the maximum value in the feature map. The resulting attention mask $A^{l'}$ is then multiplied by the original convolutional volume $D^l$.

\begin{figure}
\includegraphics[width=\textwidth]{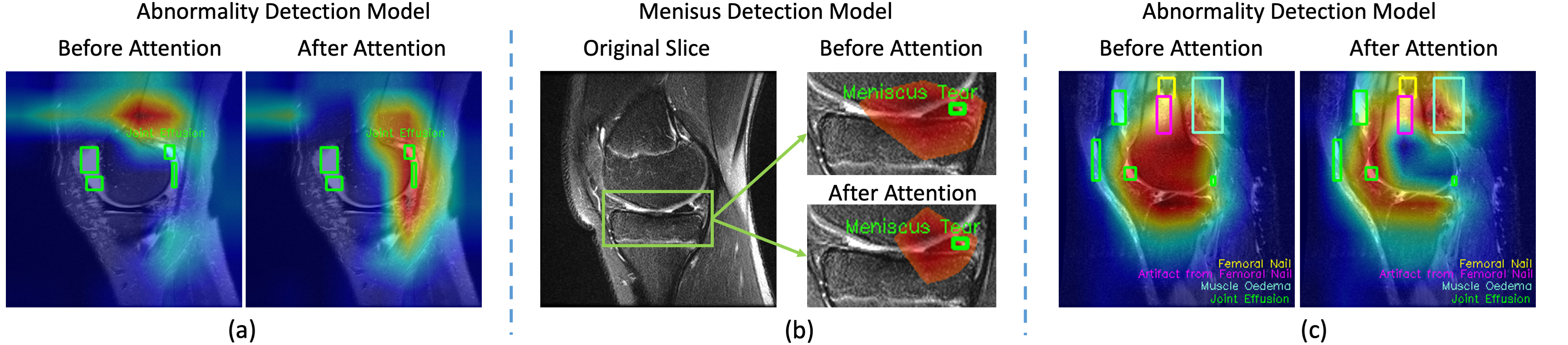}
\caption{The examples shown were generated from the same model before and after the spatial attention block was applied. (a) Spatial attention redirects the attention of the model to an abnormality (posterior joint effusion). Note that Grad-Cam overlaps with the annotations on the anterior knee in other slices. (b) A threshold is applied to Grad-Cam on both images to only colour-code the most influential regions. This then highlights how the spatial attention fine-tunes the localisation ability of the model. (c) Spatial attention removes focus from the healthy femur to focus on abnormalities. \vspace{-0.4cm}} \label{fig:att}
\end{figure}

The volume $A^{l'}$ is equivalent to a 3D spatial weight matrix where each 2D feature map in volume $A^{l'}$ is an attention mask for each feature map in volume $D^l$. Attention masks are known to switch pixels in the original volume \lq{}on\rq{} if they are important and \lq{}off\rq{} if they are not important to the target task \citep{mader}. Standard attention mechanisms develop a single 2D attention map, while our method implements a 3D attention mechanism that creates a tailored attention mask for each feature map. Figure \ref{fig:att} uses Grad-Cam to illustrate three example cases where the addition of the spatial attention block correctly redirects the model to more precisely locate a knee injury and associated abnormalities. The examples shown are MRI slices from the sagittal plane that were annotated by a radiologist on our team (I.W.).

\vspace{-0.3cm}
\subsection{Single-Plane and Multi-Plane Analysis}
\begin{figure}[htbp]
  {\includegraphics[width=1\linewidth]{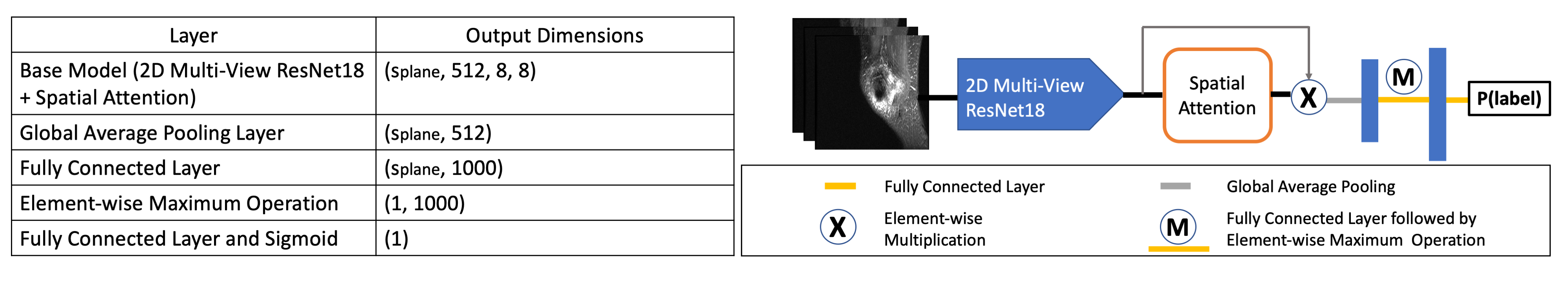}}
\vspace{-0.6cm}\caption{Output dimensions of each layer and architecture of the Single-Plane model.}
\label{fig:SP}
\end{figure}

\vspace{-0.2cm}
For the single-plane analysis, a separate model was trained for each plane and each task. Figure \ref{fig:SP} shows the architecture of the single-plane model. All slices from one plane of one MRI are input into a 2D multi-view ResNet18 with a spatial attention block of output size $(s_{plane}, 512, 8, 8)$, where $s_{plane}$ is the number of slices of an MRI from one plane. The model then has a Global Average Pooling layer of output size $(s_{plane}, 512)$, a fully connected layer of output size $(s_{plane}, 1000)$, an element-wise maximum operation of output size $(1, 1000)$, a final fully connected layer and a sigmoid function to get the final output.
\vspace{-0.4cm}

\begin{figure}[htbp]
  {\includegraphics[width=1\linewidth]{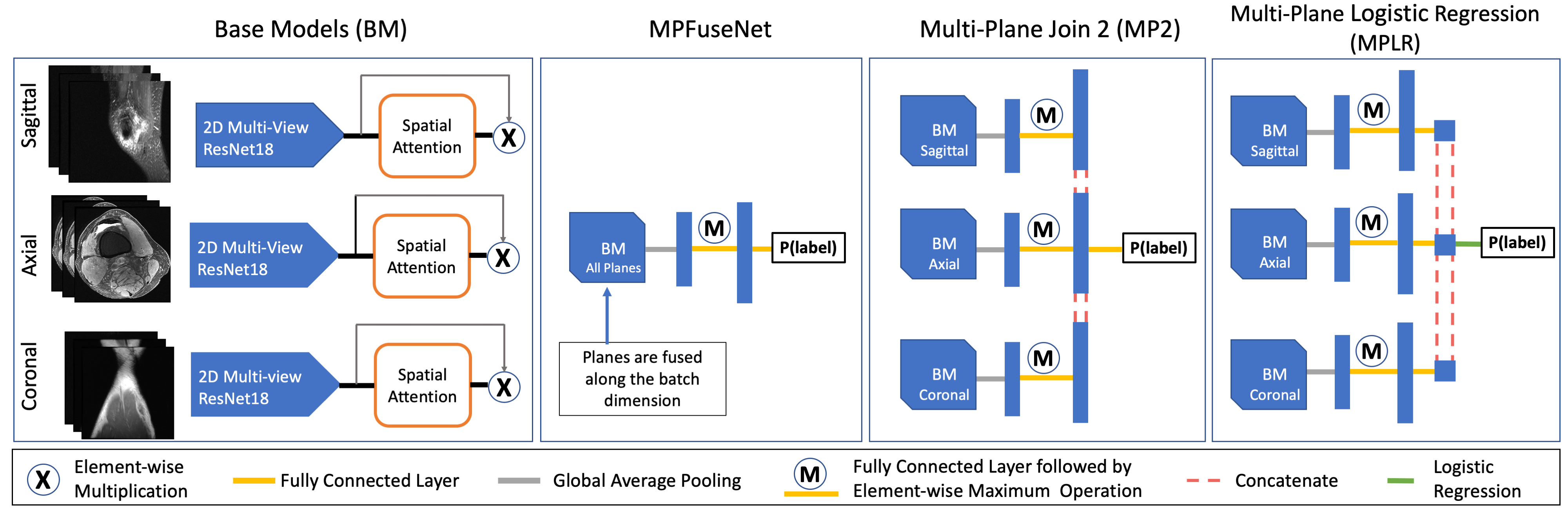} }
\vspace{-0.5cm}\caption{The BM for each plane is a 2D multi-view CNN with ResNet18 architecture and integrated spatial attention. MPFuseNet fuses the planes after the BM. MP2 fuses planes after the first fully connected layer and element-wise maximum operation. MPLR fuses the final prediction from each plane. \vspace{-.4cm}}
\label{fig:arc}
\end{figure}

\vspace{-0.4cm}
A multi-plane analysis was then conducted in order to investigate if using data from multiple planes increases injury detection ability. Figure \ref{fig:arc} visualises the multi-plane networks. These networks follow the same general architecture as the single-plane models. Three methods of fusing planes along this architecture were investigated. All of the multi-plane networks start with the Base Model (BM). The BM is the 2D multi-view ResNet18 with spatial attention block. The methods of fusing planes are described as follows.

\vspace{-0.1cm}
\renewcommand{\labelenumii}{\Roman{enumii}}
 \begin{enumerate}
 
 \item The first multi-plane join is named MPFuseNet. This network fuses the output from the BM of each plane. The output volumes from each plane are of dimensions $(s_{axial}, 512, 8, 8)$, $(s_{coronal}, 512, 8, 8)$ and $(s_{sagittal}, 512, 8, 8)$. These volumes are fused along the batch dimension, resulting in a fused volume of dimensions $((s_{axial} + s_{coronal} + s_{sagittal}), 512, 8, 8)$. 
 \item \lq{}Multi-plane Join 2\rq{} (MP2) fuses planes after the first fully connected layer and element-wise maximum operation. This converts the volume of each plane from dimensions $(1, 1000)$ to a combined volume of dimensions $(1, 3000)$. 
 \item \lq Multi-Plane Logistic Regression\rq{} (MPLR) trains each CNN separately and combines the predictions with a logistic regression model. This is the most common method in the literature \citep{mrnet, dcu}. 
 
 \end{enumerate}
 
\vspace{-0.5cm}

\vspace{-0.4cm}
\subsection{Training Pipeline}
\vspace{-0.2cm}
The models were trained on the official MRNet training set using eight-fold cross-validation and the official MRNet validation set was used as an unseen test set. Affine data augmentation techniques such as flipping, translating and rotating were applied to slices to improve model robustness. We employed the Adam optimiser \citep{adam}, with an initial learning rate of 1e-5 and a weighted cross entropy loss function. As the element-wise maximum operation for the multi-view technique is implemented over the batch dimension, the batch size is set to one to avoid combining slices of different MRIs.

\vspace{-0.2cm}
\section{Evaluation}
\vspace{-0.3cm}
\subsection{Quantitative}
\label{quantitative}
\vspace{-0.2cm}
Table \ref{tab:table1} reports the Area Under the Curve (AUC) of all multi-plane models and the best performing single-plane models for each task on the validation data. The results demonstrate that although single-plane methods can accurately detect knee injuries, including data from additional planes further improves performance. However, the training time is three times longer for multi-plane models with the number of trainable parameters tripling in comparison to single-plane models. The trade-off between performance and model complexity should be considered.

In the case of the multi-plane methods, MPLR resulted in optimal performance for detecting abnormal MRIs. However, MPLR was detrimental to performance for detecting ACL and meniscus tears. This is because single-plane models that had a lower validation AUC had a higher training AUC than other models. This resulted in the logistic regression weighting the sub-optimal performing single-plane models higher than the best performing single-plane models, resulting in a performance degradation. 

MPFuseNet performed optimally for detecting ACL and meniscus tears. This network reduces the risk of overfitting by performing the element-wise maximum operation over all planes. It does not perform optimally for abnormality detection. However, the nature of this task is substantially different to the other tasks. The models for ACL and meniscus tear detection determine whether a tear is present or not, while the abnormality detection model determines if there is a major abnormality or several minor abnormalities. Particularly in cases where the model has detected several minor abnormalities, it is possible that performing the maximum operation over all planes in MPFuseNet causes significant information loss and therefore, degrades the model's performance.

This analysis has demonstrated that MPFuseNet outperforms MPLR for ACL and meniscus tear detection and that MPLR can be detrimental to the model's performance. This is a significant finding given that MPLR is the most common method in the literature. 

From the validation results in Table \ref{tab:table1}, we select the novel MPFuseNet for ACL and meniscus tear detection and MPLR for abnormality detection as our final models.

\vspace{-0.2cm}
\begin{table}[htbp]
  
\centering
  {\caption{Comparison of Single-Plane and Multi-Plane on Validation Data (AUC)}\label{tab:table1}}%
  {\begin{tabular}{l|l|l|l}
   \textbf{Model} & \textbf{ACL} & \textbf{Abnormal} & \textbf{Meniscus} \\ \hline

\hline
 Single-Plane (SP) & 0.953$_{Axial}$ & 0.923$_{Coronal}$ & 0.862$_{Coronal}$ \\
  Multi-Plane Logistic Regression (MPLR) & 0.923 & \textbf{0.952} & 0.862 \\
  MPFuseNet & \textbf{0.972} & 0.916 & \textbf{0.867}\\
 Multi-Plane Join 2 (MP2) & 0.948 & 0.903 & 0.853 \\ 
  \end{tabular}}
\end{table}

\vspace{-0.3cm}
\cite{elnet} trained the ELNet using four fold cross-validation. Table \ref{tab:table2} compares the performance of our selected ResNet18 + Spatial Attention models and the performance of ELNet and MRNet on unseen test data, as reported by \cite{elnet}. The ELNet improved on the MRNet performance by 0.004 and 0.005 AUC for ACL and abnormality detection respectively, while our proposed models for ACL and abnormality detection result in a substantial performance improvement. Although our proposed model for meniscus detection does not outperform the ELNet, it shows an improvement on the MRNet performance.

\begin{table}[htbp]
 \centering
  {\caption{Comparison with Known Methods on Unseen Test Data (AUC)} \label{tab:table2}}%
  {\begin{tabular}{l|l|l|l}
\textbf{Model} & \textbf{ACL} & \textbf{Abnormal}& \textbf{Meniscus}  \\ \hline
 \textbf{Proposed Models} & \textbf{0.977}$_{MPFuseNet}$ & \textbf{0.957}$_{MPLR}$ & 0.831$_{MPFuseNet}$ \\
  ELNet \citep{elnet} & 0.960 & 0.941 & \textbf{0.904} \\ 
  MRNet \citep{elnet,mrnet} & 0.956 & 0.936 & 0.826 \\ 

  \end{tabular}}
\end{table}

\vspace{-0.4cm}
\subsection{Ablation Study}
\begin{figure}[htbp]

    {\includegraphics[width=1\linewidth]{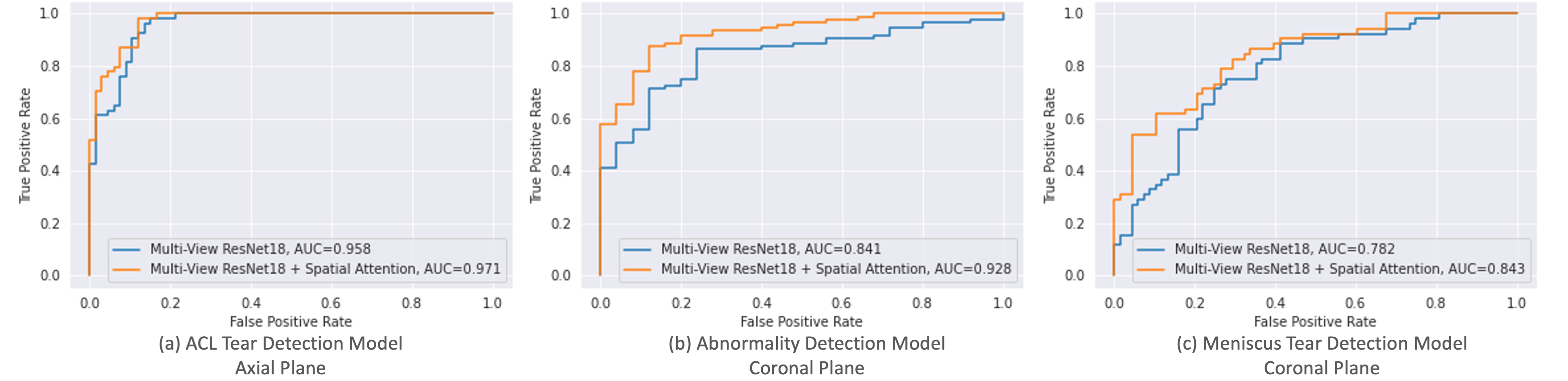}}
 \vspace{-0.6cm} {\caption{ROC curves for ResNet18 with and without spatial attention. \vspace{-0.9cm}}\label{fig:ablation}}
  
 \vspace{-0.4cm}
\end{figure}

\vspace{-0.4cm}
An ablation study was conducted to demonstrate the effects of adding the spatial attention block. A multi-view ResNet18 with no spatial attention was trained for this study. Figure \ref{fig:ablation} shows the Receiver Operator Curves (ROC) and the AUC for single-plane models with and without spatial attention on the test set. The addition of the spatial attention block results in increased performance, most notably for meniscus tear and abnormality detection.

\vspace{-0.9cm}
\section{Explainability}
\vspace{-0.2cm}
This section aims to validate the model by assessing the localisation ability and extracting features. This will aid the explainability of the model.

For this study, the radiologist annotated MRIs from 50 randomly sampled healthy and unhealthy MRIs. The radiologist used bounding boxes to annotate meniscus tears, ACL tears and significant  abnormalities  that  can  be  associative  or  non-associative of the aforementioned tears. Examples of the annotated abnormalities include joint effusion and bone bruising. Out of the sample of annotated MRIs, MRIs where the radiologist annotated an ACL tear and the single-plane model correctly detected the ACL tear were used for this analysis, resulting in a sample of twelve MRIs. There were additional cases where an ACL tear was found by the radiologist on another plane but they were not visible to the radiologist on the plane in question due to the thickness of the slices. In some of these cases, the models confidently detected a tear on the plane where it was not visible. This indicates that there are other features that the model relies on or the model has found evidence of a tear that is not visible to the human eye.

\vspace{-0.3cm}
\subsection{Localisation Ability} 
\vspace{-0.2cm}
As previously outlined, object detection models can be trained specifically to localise the tear. However, these models require localised annotations. The MRNet data set provides only image-level labels. Therefore, we use the output of Grad-Cam and the sample of MRIs that were annotated by the radiologist to quantify the model's localisation ability. This task requires a metric that meets the following two criteria. 

\noindent\textbf{Does not over-penalise false positive regions:} Grad-Cam can highlight excess area around the region of interest. This excess area outside the bounds of an annotation is a false positive region. This results in low IoU and Dice scores that over-penalise the model's localisation abilities. Furthermore, heavy penalisation of false positive regions is not ideal for assessing localisation ability for medical imaging tasks as it is likely that the area outside the bounds of the annotation may also be abnormal. For example, there could be visible joint effusion or tissue damage in the region around an ACL tear annotation. 

\noindent\textbf{Stand-alone:} Although IoU and Dice over-penalise the model's localisation abilities, they are still informative when comparing localisation abilities across models. However, a stand-alone metric can communicate an accurate representation of the localisation capabilities in the absence of any comparative information. 

A metric that meets the two criteria will accurately reflect a model's localisation ability. Our proposed metric, $PLA$, meets the aforementioned criteria. 

\vspace{-0.2cm}
\subsubsection{Localisation Accuracy and Penalised Localisation Accuracy}
\begin{figure}[htbp]

  {\includegraphics[width=1\linewidth]{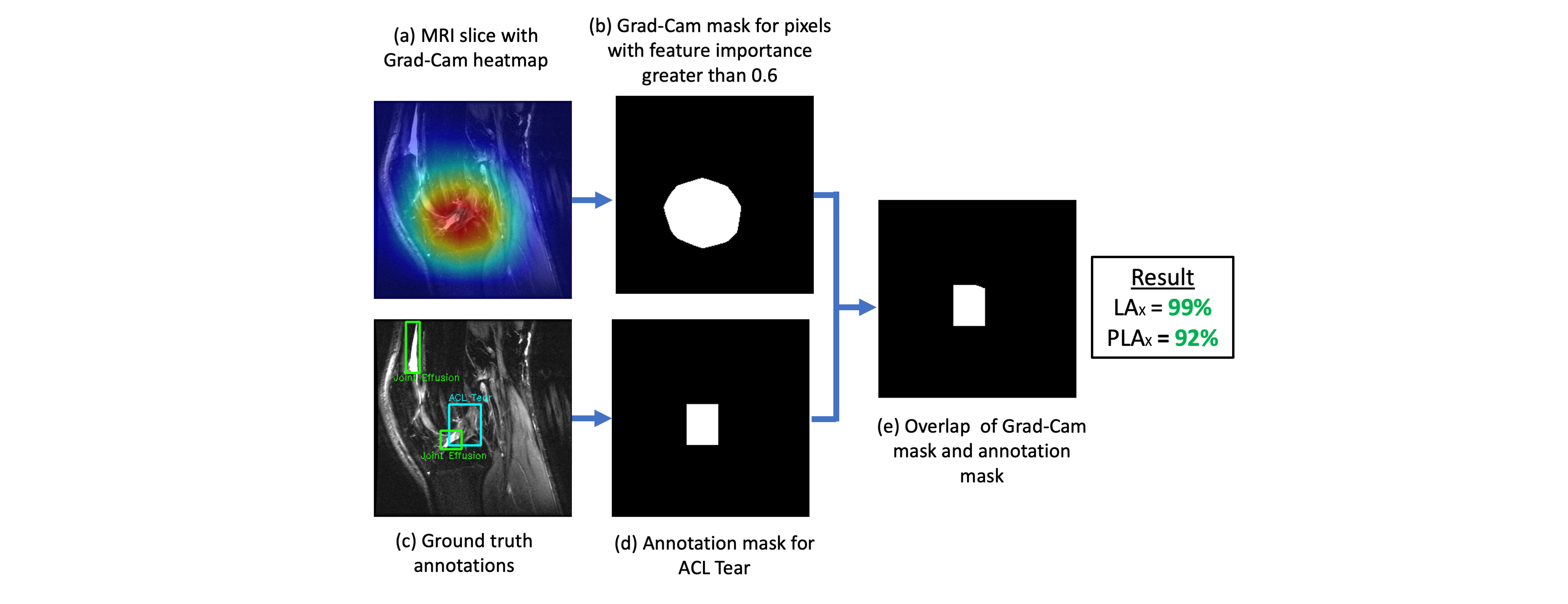}}
\vspace{-0.7cm}\caption{Pipeline for calculating $LA$ and $PLA$ based on an ACL tear detection model. 
 \label{fig:pla_pipeline} }

\end{figure}

In order to calculate Localisation Accuracy ($LA$) and $PLA$, we compare the output of Grad-Cam to the annotations defined by the radiologist. Grad-Cam generates a matrix that assigns a feature importance score between zero and one to each pixel. This matrix is normally shown as a heat-map overlaid on the MRI slice in question (figure \ref{fig:pla_pipeline}(a)). A Grad-Cam mask was generated based the pixel importance values. The masks had pixel values of one when the pixel importance value was above the threshold of 0.6 and had pixel values of zero when the pixel importance value was below the threshold of 0.6 (figure \ref{fig:pla_pipeline}(b)). The threshold of 0.6 was chosen as pixel importance values above 0.5 have above average importance. However, 0.5 was found to generate large Grad-Cam masks and higher threshold values created Grad-Cam masks that were sometimes smaller than the annotation. Therefore, the optimal threshold value is 0.6 for this study. Similarly, an annotation mask was created for each annotated MRI slice where pixels within the bounds of the annotation have a value of one and pixels outside the bounds of the annotation have a value of zero (figure \ref{fig:pla_pipeline}(d)). The Grad-Cam and annotation mask can then be used to calculate metrics such as $LA$ to quantify the localisation ability. Figure \ref{fig:pla_pipeline} outlines this process for assessing ACL tear localisation. The metric $LA$ is calculated as the percentage of the annotation that is covered by the Grad-Cam mask. The formula for $LA$ is shown in Equation \ref{eq1}. One of the limitations of this metric is that if the Grad-Cam mask covers a significantly large area, the annotation is likely to overlap with the Grad-Cam mask due to random chance. We therefore introduce an adjusted version of $LA$, $PLA$. This metric is equal to $LA$ with a penalty for false positive regions. However, unlike IoU and Dice, it does not over-penalise false positive regions. It is calculated using the formulae shown in Equation 2 and 3.

\begin{equation}
\label{eq1}
LA_{x} = \dfrac{overlap_{x}}{ann}
\end{equation}

\hspace{3.6cm}$FPP_{x} = \dfrac{gc_{x} - overlap_{x}}{total}$ \hspace{3.6cm} $(2)$

\vspace{0.35cm}

\hspace{3.2cm}$PLA_{x} = MAX(LA_{x} - FPP_{x}, 0)  \hspace{3cm}(3)$

\vspace{0.5cm}
$overlap_{x}$ is the number of pixels that are both, within the bounds of the annotation and have a pixel value of one in the Grad-Cam mask. The subscript $x$ means that the metric was calculated based on a Grad-Cam mask generated at pixel importance threshold $x$. As outlined earlier, we set $x$ to equal the pixel importance threshold 0.6. $ann$ is the number of pixels within the bounds of the radiologist's annotation. $FPP_{x}$ is the False Positive Penalty (FPP). $gc_{x}$ is the number of pixels that have a value of one in the Grad-Cam mask and $total$ is the total number of pixels in the image. Equation 3 also shows a maximum operation. This is to avoid negative $PLA$ values. Without a maximum operation, $PLA$ would have a negative value when there is no overlap between the Grad-Cam mask and the annotation mask.

\vspace{-0.6cm}
\vspace{-0.55cm}
$PLA$ returns a value between zero and one where a score of one is perfect localisation. A score of one is achieved when the area of the Grad-Cam mask is equal to the area of the annotation mask and they have a perfect overlap. A score of zero is achieved if there is no overlap between the Grad-Cam mask and the annotation mask. $PLA$ also effectively penalises false positive regions. For example, if the Grad-Cam mask covers 100\% of the image and the annotation takes up 1\% of the area, the value of $LA$ would be one, while the value of $PLA$ would be 0.01. Although $PLA$ penalises false positive regions, it does not over-penalise them as is the case with IoU and Dice. This is demonstrated in a later paragraph. $PLA$ is also a stand-alone metric, meaning it is interpretable and informative of the localisation ability without any comparative information. 

\vspace{-0.35cm}
\subsubsection{Area Under the Curve for Localisation}
The Area Under the Curve is another metric that can be used to assess a model's localisation ability that meets the previously outlined criteria. To calculate the AUC, the Grad-Cam output and annotation masks are flattened into 1D vectors. The Grad-Cam 1D vector is interpreted as the probability of the pixel showing the tear and the annotation 1D vector is the ground truth. The AUC is then calculated. This metric takes into account false positive regions and is advantageous over IoU and $PLA$ as it does not require a Grad-Cam mask to be generated at a specific threshold.  Instead, the direct Grad-Cam output can be used for the AUC calculation. However, it is not as easily interpreted as our proposed metric, $PLA$. 

\vspace{-0.2cm}
\subsubsection{Metric Comparison}

\begin{figure}[htbp]

  {\includegraphics[width=1\linewidth]{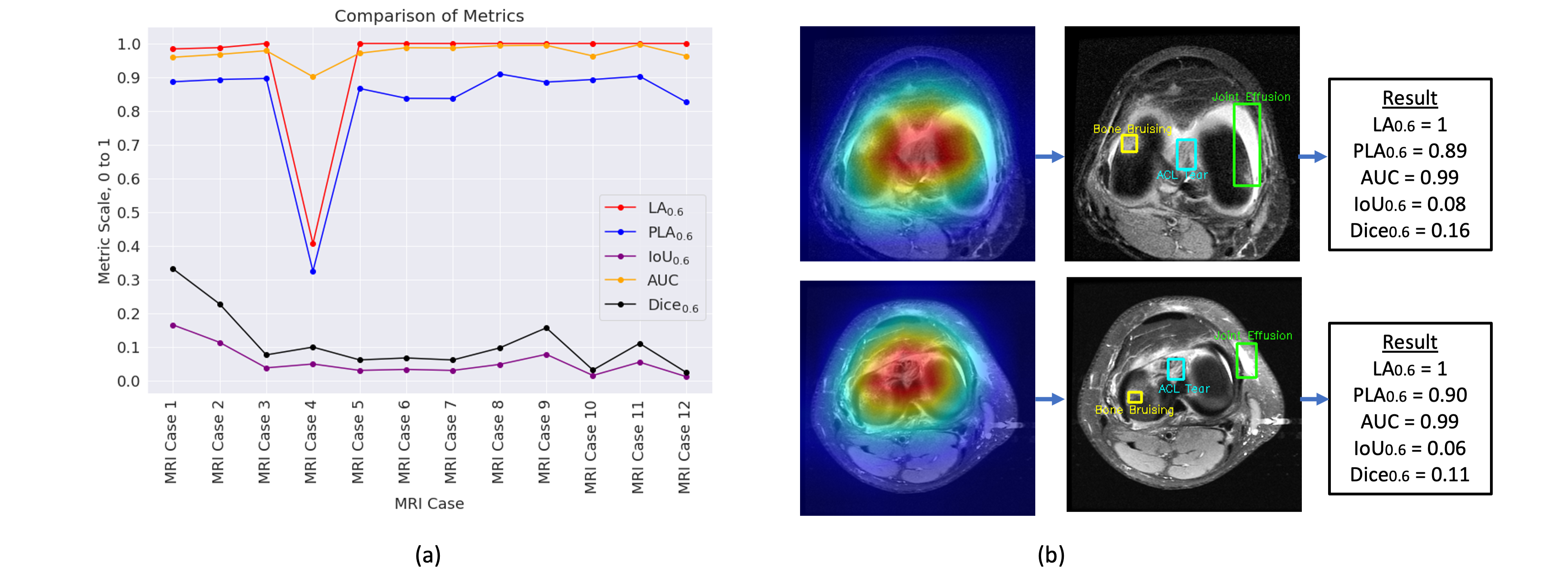}}
\vspace{-0.6cm}\caption{(a) The $LA_{0.6}$, $PLA_{0.6}$, AUC, IoU$_{0.6}$ and Dice$_{0.6}$ scores for sample MRIs. (b) Grad-Cam overlaid on axial plane MRI slices and the corresponding radiologist annotations for MRI Case nine (top) and eleven (bottom). These metrics are based on the model's ability to locate the ACL tear (ACL tear annotations are shown in blue).
 \label{fig:local_results} }
\vspace{-0.6cm}
\end{figure}

Figure \ref{fig:local_results}(a) shows the value of each metric on a sample of MRIs. These scores are based on the ACL detection model that was trained on axial data. For each MRI, the metrics shown in the figure are based on the slice with the highest metric values as we interpret this to be the MRI's key slice. The metrics are shown on the same graph as all metrics have a range of possible values from zero to one. All metrics, with the exception of AUC, were calculated using a Grad-Cam mask generated at the 0.6 pixel importance threshold. AUC uses the direct Grad-Cam output. Figure \ref{fig:local_results}(b) shows the Grad-Cam heatmap overlaid on the MRI slice and the corresponding radiologist annotations. As we are assessing the model's ability to localise an ACL tear, it is only the ACL tear annotations shown in blue that are of interest. Figure \ref{fig:local_results}(b) also shows the values of the metrics for each example. The radiologist was satisfied with the model's localisation of the ACL tear in the examples shown. However, the IoU and Dice metrics have near zero values, indicating poor localisation ability. It can be seen from Figure \ref{fig:local_results}(a) that all samples have low IoU and Dice scores. The disagreement between these metrics and the radiologist's opinion demonstrates the ineffectiveness of IoU and Dice as stand-alone, interpretable metrics that can be used to assess localisation ability. This is due to their over-penalisation of false positive regions. A further cause for concern is that neither IoU or Dice show a dis-improvement in their scores for the problematic MRI Case 4. In this MRI case, the Grad-Cam mask overlaps with only 40\% of the ACL tear annotation. 
\vspace{-0.4cm}

Figure \ref{fig:local_results}(b) shows that $LA$ achieves a perfect score for almost all samples. However, $LA$ will not accurately communicate the localisation ability of a model in cases where there are significantly large false positive regions. It can be concluded from the figure that our proposed $PLA$ and AUC give the most accurate representation of the model's localisation ability. They both give high scores when the model has effectively localised the tear but still penalise false positive regions. An advantage of AUC is that it does not require a Grad-Cam mask to be generated based on a pixel importance threshold. It uses the direct Grad-Cam output. However, it appears to under-penalise the model in cases. MRI Case four, as previously outlined, only overlaps with 40\% of the annotation. However, this case achieves a high AUC score of 0.9. Therefore, $PLA$ can be considered to be advantageous over AUC. Moreover, our proposed metric, $PLA$ is akin to a straight-forward accuracy calculation with an adjustment for false positive regions and therefore, it is more interpretable than AUC.

\vspace{-0.2cm}
\subsubsection{Aggregated Localisation Results}
To quantify the localisation ability over the entire sample, we can calculate the percentage of cases where the model accurately located the tear based on the $PLA$ scores. Value a$^s_{x}$ is the $PLA$ value of slice $s$ of an MRI that overlaps with the Grad-Cam mask segmented at feature importance threshold $x$ where $x$ is equal to 0.6 for this study. The slice with the best a$^s_{x}$ score is selected from each MRI as we interpret this slice to be the MRI's key slice. For each MRI, a$^s_{x}$ is subsequently transformed to a binary outcome of [0, 1] using a threshold value $k$. The value $k$ can be varied in the range (0.5, 1). For each value of k, a$^s_{x}$ is assigned a value of one if k is less than a$^s_{x}$ and zero otherwise. Once a binary outcome has been obtained for each MRI, the accuracy can be calculated over the sample for each threshold value $k$. We report that 91.7\% of the sample accurately localised the tear at $k$ values ranging from 0.5 to 0.85. It is evident that our proposed model is correctly localising ACL tears and the output of the model is based on the site of the ACL tear. Moreover, all MRI cases have an AUC greater than 0.9. This further demonstrates the model's localisation ability. 

\vspace{-0.3cm}
\subsection{Features}
\vspace{-0.2cm}
The radiologist's annotated abnormalities were used to extract features. The purpose of annotating abnormalities in addition to the ligament tears is to determine if there are additional structures present in the MRI that influence whether a model detects a ligament tear or not. Such \lq structures\rq{} can be thought of as features. Figure \ref{fig:feat} visualises the most common abnormalities that the ACL tear model detected. The table shows the number of MRIs where the feature was present and the detection rate. The detection rate is calculated based on number of times the Grad-Cam mask generated at the 0.6 pixel importance threshold covered over 60\% of the annotation.  The results show that joint effusion and bone bruising are the most common features of an ACL tear. The radiologist verified that these features are frequently seen in the setting of acute ACL tears.

It was also found from an analysis of the ACL detection model trained on sagittal data that there was a correlation between abnormal cases and cases where growth plates were visible on the MRI. Growth plates are found in children and adolescents. Grad-Cam highlighted these normal regions to indicate that they are influential in the model's output. Therefore, the model learned an incorrect relationship. This is an example of the model falling victim to the correlation versus causation fallacy. This further justifies the requirement of extracting features and clinically verifying them. 


\vspace{-0.3cm}

\subsection{Limitations}
\begin{figure}[htbp]
  {\includegraphics[width=1\linewidth]{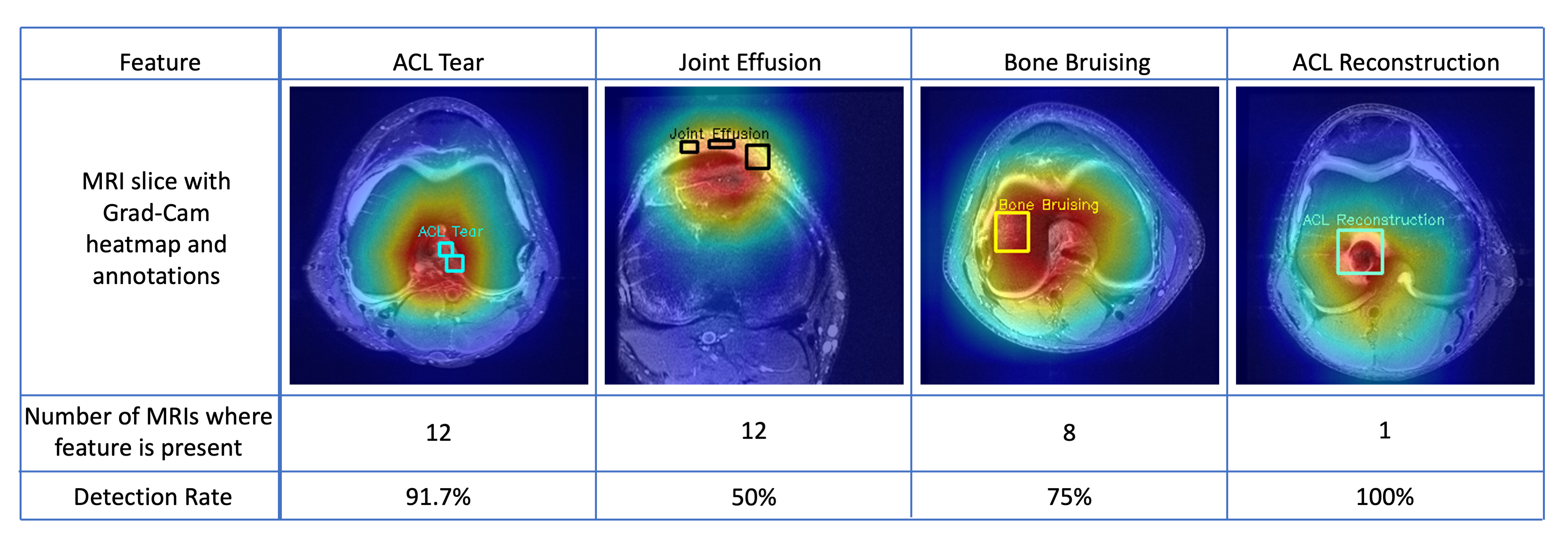}}
\vspace{-0.8cm}\caption{Table demonstrating the number of MRIs where the feature is present and the feature detection rate.
 \label{fig:feat} }
\vspace{-0.3cm}
\end{figure}

\vspace{-0.2cm}
Although the experiment quantified the localisation ability of the model and extracted features, there were a number of limitations. The annotations were limited to bounding boxes. This could result in excess area being included in the annotation if the abnormality is not rectangular. The radiologist minimised the effects of this by using multiple annotations for irregularly shaped abnormalities. An example of this is the joint effusion annotation in Figure \ref{fig:feat}. The radiologist also noted that the slice thickness of some MR protocols made it difficult to detect and annotate tears. Grad-Cam has some known limitations such as not capturing the entire region of interest \citep{gradplus}. A variation of visualisation techniques could be employed in future to overcome this limitation.

\vspace{-0.3cm}

\section{Conclusion}
\vspace{-0.2cm}
We have demonstrated how spatial attention corrects the model's focus to salient regions. In our analysis of single and multi-plane, it was found that including data from additional planes increases the detection ability. However, the training time for multi-plane methods and the number of parameters is tripled and thus, the trade-off between model complexity and performance should be considered. We developed the multi-plane network, MPFuseNet, that outperforms the common MPLR method for ACL and meniscus tear detection. Furthermore, our proposed methods achieve state-of-the-art results for ACL tear and abnormality detection. 

We developed an objective metric, $PLA$ for quantifying and validating the localisation ability of our model. Using this metric, we verified that the ACL detection model accurately located the ACL tear in 91.7\% of a sample of MRIs.  We then extracted features from the model that were then verified by the radiologist. Validating the localisation ability and extracting features for explainability will improve clinical trust in ML systems and allow for smoother integration into clinical workflows. This work will be extended to consider the opinion of several radiologists and a larger data set of MRIs. Future work will also elaborate on the single-plane and multi-plane analysis by considering additional planes. Oblique MRI planes have previously been shown to improve ACL tear detection \citep{kosaka}. The quality of the extracted explainability features will also be assessed using the System Causability Scale \citep{holzinger} and by means of a user evaluation study with clinicians. 

Code for this work will be made publicly available at https://git.io/J3pA5. 

\vspace{-0.4cm}
\section*{Acknowledgements}
\vspace{-0.3cm}
This work was funded by Science Foundation Ireland through the SFI Centre for Research Training in Machine Learning (18/CRT/6183).
This work is supported by the Insight Centre for Data Analytics under Grant Number SFI/12/RC/2289\_P2. This work was funded by Enterprise Ireland Commercialisation Fund under Grant Number CF 201912481. We would also like to acknowledge and thank Professor Kevin McGuinness from Dublin City University (DCU) for his guidance on the research study.

\bibliography{ref}

\end{document}